\documentclass{article}

\usepackage{float}
\usepackage{fancyvrb}
\usepackage{fvextra}
\usepackage{xurl}

\usepackage{microtype}
\usepackage{graphicx}
\usepackage{subcaption}
\usepackage{booktabs} 

\usepackage{hyperref}

\usepackage[accepted]{icml2026}

\usepackage{amsmath}
\usepackage{amssymb}
\usepackage{mathtools}
\usepackage{amsthm}

\usepackage[capitalize,noabbrev]{cleveref}

\theoremstyle{plain}

\theoremstyle{definition}

\theoremstyle{remark}

\usepackage[textsize=tiny]{todonotes}

\icmltitlerunning{Representations of Text and Images Align From Layer One}

\begin{document}

\twocolumn[
\icmltitle{Representations of Text and Images Align From Layer One}



\icmlsetsymbol{equal}{*}

\begin{icmlauthorlist}
\icmlauthor{Evžen Wybitul}{eth}
\icmlauthor{Javier Rando}{eth}
\icmlauthor{Florian Tramèr}{eth}
\icmlauthor{Stanislav Fort}{aisle}
\end{icmlauthorlist}

\icmlaffiliation{eth}{D-INFK, ETH Zurich, Switzerland}
\icmlaffiliation{aisle}{Aisle Research}
\icmlcorrespondingauthor{Evžen Wybitul}{evzen@robots.ox.ac.uk}

\icmlkeywords{Machine Learning, ICML}

\vskip 0.3in
]



\printAffiliationsAndNotice{} 

\begin{abstract}
We show that for a variety of concepts in adapter-based vision-language models, the representations of their images and their text descriptions are meaningfully aligned from the very first layer.
This contradicts the established view that such image-text alignment only appears in late layers.
We show this using a new synthesis-based method inspired by DeepDream: given a textual concept such as ``Jupiter'', we extract its concept vector at a given layer, and then use optimisation to synthesise an image whose representation aligns with that vector.
We apply our approach to hundreds of concepts across seven layers in Gemma 3, and find that the synthesised images often depict salient visual features of the targeted textual concepts: for example, already at layer 1, more than 50~\% of images depict recognisable features of animals, activities, or seasons.
Our method thus provides direct, constructive evidence of image-text alignment on a concept-by-concept and layer-by-layer basis.
Unlike previous methods for measuring multimodal alignment, our approach is simple, fast, and does not require auxiliary models or datasets. 
It also offers a new path towards model interpretability, by providing a way to visualise a model's representation space by backtracing through its image processing components.
\end{abstract}

\section{Introduction}

\begin{figure}[ht!]
\centering
\includegraphics[width=0.99\columnwidth]{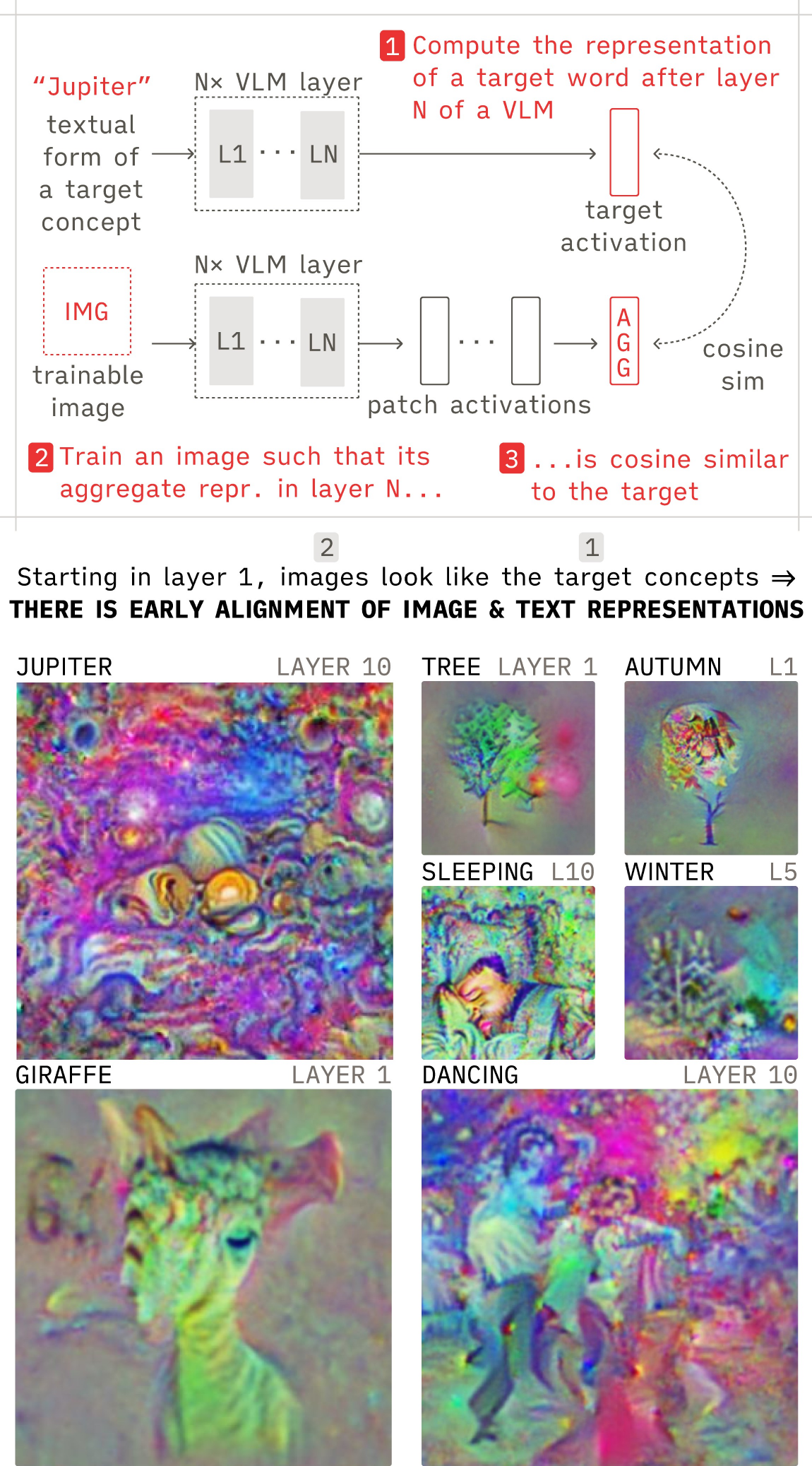}
\caption{\textbf{Top:} Given a text description of a concept (e.g., ``Jupiter'') and a target layer in a vision-language model, our method synthesises an image whose representation aligns with that of the target word in that layer.
This enables studying how the models bridge between the modalities.
\textbf{Bottom:} Surprisingly, as early as layer 1, the synthesised images are often recognisable as depicting the target concepts, demonstrating that representations of text and images align much earlier than previously thought.}
\label{fig:main}
\end{figure}

Understanding how vision-language models (VLMs) bridge between text and images is fundamental to interpreting their behaviour. 
In adapter-based VLMs, which connect a pre-trained vision encoder to a language model, a key question is when and how the representations of concepts in text and images align.
Previous work has found that early layers show a ``modality gap''~\cite{jiang2024hallucinationaugmentedcontrastivelearning}, and alignment between text and images only emerges in mid-to-late layers~\cite{venhoff2025visual,neo2025interpretingvisualinformationprocessing}.
In this work, we challenge this view: using a novel method based on synthesising images, we show that representations of images and text meaningfully align as early as layer 1 for many concepts.
This happens despite models never being explicitly trained to align representations at any particular layer.

Our approach differs fundamentally from the prior ones, which do not provide direct, concept-level evidence of when visual and textual representations align layer by layer.
The most comprehensive analysis to date by~\citet{venhoff2025visual} used proxy metrics.
Rather than examining specific concepts across layers, their work compared the reconstruction error and sparsity of text-based sparse autoencoders~\cite{cunningham2023sparseautoencodershighlyinterpretable} when applied to image representations.
\citet{neo2025interpretingvisualinformationprocessing} similarly found alignment in late layers --- in this case, by showing that image patches can be decoded to words describing the objects at those patch locations.
However, this approach uses the model's final-layer decoding machinery~\cite{nostalgebraist2020logitlens}, which a priori makes it less suitable for probing early layers.
Lastly, even methods designed for cross-modal analysis face fundamental constraints.
Multimodal sparse autoencoders~\cite{lou2025saevinterpretingmultimodalmodels} require lengthy training for each model and, more critically, interpret concepts by identifying which examples in existing datasets activate them --- a process that risks introducing confounders and biases when not paired with more rigorous analysis~\cite{stevens2025sparseautoencodersscientificallyrigorous}.

We take an approach that is much more direct.
Instead of searching through existing datasets, we synthesise images from scratch: we optimise images such that their representations at specific layers align with those of textual concepts, using gradient-based techniques from adversarial robustness~\cite{fort2025directascentsynthesisrevealing}.
For example, as illustrated in \cref{fig:main}, given the representation of the word ``Jupiter'' at layer 10, we synthesise an image that has a similar representation at that layer. 
Crucially, we validate the synthesised image by testing whether an independent model recognises it as depicting Jupiter; recognition implies that our optimisation discovered genuine visual-textual correspondence.
This synthesis-based approach is simple (gradient optimisation only), fast (minutes, not hours), and clean (no datasets or auxiliary models).

\textbf{Our first contribution} is the method itself: a dataset-free technique for visualising concepts in neural networks through generative synthesis.
While we apply it here to probe vision-language alignment, the core approach --- generating prototype images that serve as visual representations of textual concepts --- holds broad potential for interpretability research. 
Unlike existing methods, ours requires no auxiliary training, no external datasets, and produces directly interpretable images.

\textbf{Our second contribution} is the surprising finding that representations of text and images align much earlier than previously thought.
Applying our method to over 100 concepts across seven layers of Gemma 3 4B~\cite{team2025gemma}, we find that multiple concept categories, including animals, celebrations, and seasons, produce recognisable prototype images from layer 1.
This occurs despite Gemma never being explicitly trained to align visual and textual representations; the model is trained to process vision encoder outputs alongside text tokens for next-token prediction.
These findings establish a lower bound on early representation alignment, with future refinements of the method likely to reveal even more extensive cross-modal correspondence.

\section{Method: Generating Image Prototypes for Language Concepts} \label{section:method}

Our goal is to investigate the alignment between representations of images and text in vision-language models. 
We do this by testing whether it is possible to recover \textit{visual} information from the representation of a \textit{textual} concept.

Concretely, we synthesise an image whose representation at some target layer aligns with the representation of a textual concept at that layer.
Then, we check whether the synthesised image contains visual information about the concept by testing if the image \textit{transfers:} we show the image to a model that was not used during optimisation and verify whether the model recognises the image as depicting the target concept.
If it does, we have constructive evidence that the textual representation contained recoverable visual information --- in other words, that representations of images and text of that concept are meaningfully aligned.

\paragraph{Problem statement.} Given a textual concept $w$ (e.g., the word ``Jupiter'') and a target layer $\ell$, we seek an image $I^*$ whose representation has high cosine similarity with that of $w$ and which transfers to other models:
\begin{align*}
I^* = \operatorname{argmax}_{I} \operatorname{sim}(\operatorname{rep}_\ell(I), \operatorname{rep}_\ell(w)) \\ \text{such that } I^* \text{ transfers to other models,}
\end{align*}
where $\operatorname{rep}_\ell(I) \in \mathbb{R}^H, \operatorname{rep}_\ell(w) \in \mathbb{R}^H$ are representations of an image and a textual concept at layer $\ell$, respectively, and $H$ is the hidden dimension of the model.

This formulation involves three design decisions that we address in the following sections:
\begin{enumerate}
    \item \textbf{\cref{section:concept_directions}:} Obtaining $\operatorname{rep}_\ell(w)$, the representation of a textual concept. We use a simple method based on steering vectors.
    \item \textbf{\cref{section:patch_aggregation}:} Computing $\operatorname{rep}_\ell(I)$, the representation of an image. We aggregate over patch activations using a weighted scheme.
    \item \textbf{\cref{section:das}:} Guiding optimisation toward images that transfer. We borrow techniques from the adversarial robustness literature.
\end{enumerate}

See \cref{appendix:motivation} for a toy demonstration that motivates our use of cosine similarity, and \cref{appendix:full_loss_function} for a full formulation of the final loss function.

\subsection{Representing Textual Concepts via Steering Vectors} \label{section:concept_directions}

To obtain $\operatorname{rep}_\ell(w)$, the representation of a textual concept $w$ at layer $\ell$, we extract a \textit{concept direction} --- a vector in activation space that captures the semantic content of the concept~\cite{turner2024steeringlanguagemodelsactivation,panickssery2024steeringllama2contrastive}.

Our method is agnostic to how we obtain the concept directions.
They could come from sparse autoencoders~\cite{cunningham2023sparseautoencodershighlyinterpretable}, contrastive activation methods~\cite{turner2024steeringlanguagemodelsactivation,panickssery2024steeringllama2contrastive}, probing classifiers~\cite{belinkov2021probingclassifierspromisesshortcomings}, or other sources. 
For our experiments, we use a recent method from~\citet{lindsey2025emergent}, which we describe below.

For a target concept $w$ (typically a single word or short phrase, e.g., ``Jupiter'' or ``San Francisco''), we extract its activation and subtract a baseline to centre the activation and isolate its semantic component:
\begin{equation} \label{eq:lang_concept}
\operatorname{rep}_\ell(w) = \operatorname{activation}_\ell (w) - \mathbf{b}^\text{lang}_{\ell},
\end{equation}
where $\mathbf{b}^\text{lang}_{\ell} \in \mathbb{R}^{H}$ is the baseline vector, whose computation we describe below.
To compute $\operatorname{activation}_\ell : \text{String} \to \mathbb{R}^{H}$, we wrap $w$ in the model's chat template (as if a user said it) and extract activations directly from the token positions of $w$ in the target layer $\ell$, averaging them across tokens for multi-token $w$.

This differs from ~\citet{lindsey2025emergent} who compute the activation not on the raw $w$, but on ``Tell me about $w$,'' and extract the activation not from the word's token positions, but from the last token position, i.e., the assistant turn token.
This modification is motivated by our focus on early layers, where information about a concept resides primarily at the token positions describing it, as the model has not yet propagated information to other positions.

The centring is necessary because activations are not centred at zero; intuitively, each activation can be decomposed as the sum of a common component (shared across words) and a semantic component (word-specific).
Without centring, cosine similarities are dominated by the shared component.
The baseline $\mathbf{b}^{\text{lang}}_{\ell}$ approximates the common component, and subtracting it allows us to isolate what is semantically distinctive about each concept.

We compute the baseline  $\mathbf{b}^{\text{lang}}_{\ell}$ as the mean activation across a set of 100 baseline words $W_\text{baseline}$ spanning concrete objects (``gondola'') and abstract concepts (``knowledge''):
\begin{equation} \label{eq:lang_baseline}
\mathbf{b}^\text{lang}_{\ell} = \frac{1}{|W_{\text{baseline}}|} \sum_{w \in W_{\text{baseline}}} \operatorname{activation}_{\ell}(w),
\end{equation}
again averaging over token positions for multi-token words.
The set $W_\text{baseline}$ comes directly from~\citet{lindsey2025emergent} (converted to singular form); the complete list is in \cref{appendix:word_baseline}.

\paragraph{Relationship to traditional contrastive methods.} 
This approach to finding concept directions simplifies the traditional contrastive setup where one constructs many contrastive prompt pairs (e.g., ``I talk about weddings'' vs. ``I don't talk about weddings'') and averages their differences~\cite{turner2024steeringlanguagemodelsactivation,panickssery2024steeringllama2contrastive}.
\citet{lindsey2025emergent} found that this simpler approach still produces effective concept directions.
We adopted it to prevent inadvertent prompt-specific artefacts from confounding concept-specific signals in our analysis.
Related approaches, such as TCAV \citet{kim2018interpretabilityfeatureattributionquantitative}, also interpret concepts as directions in activation space.

\subsection{Representing Images via Weighted Patch Aggregation} \label{section:patch_aggregation}

To compute $\operatorname{rep}_\ell(I)$, we must first understand how adapter-based VLMs represent images.
These models process images through three stages: (1)~a vision encoder extracts visual features into a sequence of patch embeddings, (2)~(optionally) a linear projector maps these patch embeddings to the language model's embedding space, (3)~the resulting sequence of patch embeddings is concatenated to the sequence of text token embeddings and processed by the language model.
The vision encoder is pre-trained separately and often stays frozen during training.

After each transformer layer $\ell$ with hidden dimension $H$, an image $I$ is thus represented as a sequence of $N$ patch activations:
\begin{equation*}
\operatorname{patches}_{\ell}(I) = (\mathbf{p}_1, \mathbf{p}_2, \ldots, \mathbf{p}_N), \quad \mathbf{p}_i \in \mathbb{R}^{H}.
\end{equation*}

We define $\operatorname{rep}_\ell(I)$ as a weighted average over centred patch activations $\mathbf{p}^\text{cent}_i \in \mathbb{R}^{H}$:
\begin{equation*}
\operatorname{rep}_\ell(I) = \sum_{i=1}^N \alpha_i \mathbf{p}^\text{cent}_i,
\end{equation*}
where $\alpha_i \in \mathbb{R}$ are attention weights that concentrate on semantically relevant patches while maintaining spatial coherence.
We describe the computation of both the centred patch activations and the weights below.

We centre each patch by subtracting an image baseline:
\begin{equation*}
\mathbf{p}^\text{cent}_i = \mathbf{p}_i - \mathbf{b}^\text{img}_{\ell},
\end{equation*}
where $\mathbf{b}^\text{img}_{\ell} \in \mathbb{R}^{H}$ is the mean patch activation of a neutral grey image (pixels = 0.5):
\begin{equation*}
\mathbf{b}^\text{img}_{\ell} = \frac{1}{N} \sum_{j=1}^N \operatorname{patches}_\ell(I_\text{grey})_j.
\end{equation*}
This is analogous to centring language activations (\cref{eq:lang_baseline,eq:lang_concept}), and has the goal of isolating patch-specific semantic content.

The weights $\alpha_i$ combine semantic relevance ($s_i$) with spatial coherence ($g_i$):
\begin{equation*}
\alpha_i = \operatorname{softmax}_\tau(s_i + \log g_i),
\end{equation*}
where $\tau$ is the temperature (low values focus attention on the most relevant patches) and $i$ is the visual patch / token index.

The semantic score $s_i \in [-1, 1]$ measures how similar each centred patch is to the representation of the target concept:
\begin{equation*}
s_i = \operatorname{cosine\_similarity}(\mathbf{p}^\text{cent}_i, \operatorname{rep}_\ell(w)).
\end{equation*}

The spatial prior $g_i \in \mathbb{R}^+$ is a Gaussian centred on the image centre:
\begin{equation*}
g_i = \exp\left(-\frac{(x_i - x_c)^2 + (y_i - y_c)^2}{2\sigma^2}\right),
\end{equation*}
with $(x_i, y_i)$ the spatial coordinates of patch $i$, $(x_c, y_c)$ the centre of the image, and $\sigma$ controlling the prior's spread (see below for our choice).
This encourages the optimisation process to produce a single object in the centre of the image rather than fragmented patterns.

In our experiments, we use $\tau = 0.005$, or $\tau=0.5$, depending on the layer, and we update the $\sigma$ of the Gaussian coherency prior during training according to a linear schedule, from relatively focused ($\sigma = 2$) to diffuse ($\sigma= 16$). 
We list the remaining experimental details in \cref{section:results_experimental_setup} and \cref{appendix:results_experimental_setup}.

\subsection{Guiding Optimisation Toward Transfer via Direct Ascent Synthesis} \label{section:das}

\begin{figure}
\centering
\includegraphics[width=0.85\columnwidth]{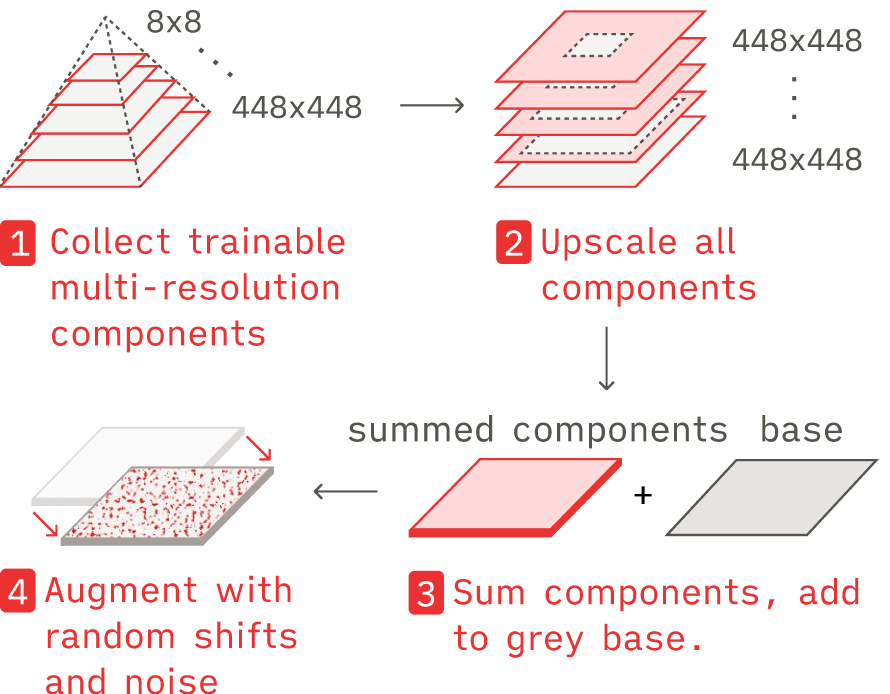}
\caption{
Our image synthesis pipeline based on direct ascent synthesis~\cite{fort2025directascentsynthesisrevealing}. 
The final image consists of a base image (neutral grey) plus a trainable perturbation. 
(1)~Instead of training the perturbation pixels directly, we re-parametrise the perturbation via a set of trainable multi-resolution components.
(2--3)~All multi-resolution components are upscaled to the target size, summed, and added to the base image.
(4)~The resulting image is augmented with random shifts and noise.
}
\label{fig:das}
\end{figure}

Unconstrained optimisation in high-dimensional pixel space can discover adversarial patterns that exploit quirks of the specific model rather than recovering genuine visual content.
These images transfer poorly.
To encourage transfer, we thus use direct ascent synthesis (DAS), a technique from the adversarial robustness literature~\cite{fort2025directascentsynthesisrevealing}.
See \cref{fig:das} for an overview.

DAS, at its core, is an adversarial attack method: it starts with a base image (in our case, neutral grey) and generates a perturbation to be added to it, optimising the resulting image to minimise some loss function (in our case, the dissimilarity between the image's representation and the target concept direction).
During optimisation, it applies random shift and noise augmentations to guide the process toward more natural-looking images.
However, the key difference is that instead of directly optimising the pixels of the perturbation, DAS re-parametrises it as a sum of components at multiple resolutions, which are then optimised simultaneously.

\paragraph{Step 1: Obtain the base image.} 
We chose a target resolution for our synthesised image to be
$$
R \times R = 448 \times 448\, ,
$$
The base image $I_{\text{base}} \in [0, 1]^{R \times R \times 3}$ is a neutral grey image with all pixel values set to $0.5$.

\paragraph{Step 2: Compute the perturbation as a sum of multi-resolution components.} 
We do not optimise the perturbation pixels directly.
Instead, we maintain a set of separate trainable components at a range of resolutions,
\[
\{\text{layer}_{r \times r} \in \mathbb{R}^{r \times r \times 3} \} \text{ for } r \in \{R, R-20, R-40,\ldots\},
\]
with the smallest resolution being $8\times 8$.
We construct the final perturbation by upscaling each component to the target resolution and summing them:
\[
\text{perturbation} = \sum_{r} \text{upscale}_{R \times R}\left(\text{layer}_{r \times r}\right).
\]
We scale this sum with $\tanh$ and add it to the base image:
\[
I_{\text{final}} = I_{\text{base}} + \frac{1}{2} \tanh \left( \text{perturbation} \right).
\]
Due to the $\tanh$ scaling and $I_{\text{base}}$ being a neutral grey image, $I_{\text{final}}$ is guaranteed to be in $[0, 1]$ and thus a valid image.

This multi-resolution reparameterisation encourages the optimisation to capture both coarse semantic structure (in low-resolution components) and fine details (in high-resolution components) simultaneously.

\paragraph{Step 3: Augment the image.} 
Before passing the image to the VLM, we apply random augmentations.

We shift the image horizontally and vertically by an amount sampled uniformly from $\{-\Sigma_\text{DAS}, \ldots, \Sigma_\text{DAS}\}$ pixels.
We first upscale the image $R + \Sigma_\text{DAS}$ pixels in each direction, apply the shift, then centre-crop back to the target resolution $R \times R$. 
We then add Gaussian noise with standard deviation $\sigma_\text{DAS}$ to each pixel.

Following the original DAS paper, we use $\Sigma_\text{DAS} = 56$ and $\sigma_\text{DAS} = 0.1$, and when optimising in batches, we apply the augmentations independently to each batch element.

\paragraph{Transfer to other models and recognisability by humans.}
Our approach to regularisation involves trade-offs.
For our current application --- testing whether alignment exists --- getting transfer is sufficient: if an independent model recognises the synthesised image, we have evidence of alignment.
However, potential future applications of this method as an interpretability tool --- a visual lens into a model's representation space --- would require not just transfer but \textit{human} recognisability, which is a stronger requirement.
Our current method already produces human-recognisable images for many concepts (\cref{section:results}), suggesting it is partway toward this goal, but does not do so reliably.
On the other hand, the regularisation techniques we use may already sometimes be too restrictive, causing us to miss genuine alignment that does not manifest in the forms that DAS-regularised optimisation favours.
Adjusting the strength and type of the transfer-inducing techniques is thus a natural direction for future work.

\section{Results: Early Vision-Language Alignment}\label{section:results}

\begin{figure*}[t]
\centering
\includegraphics[width=0.99\textwidth]{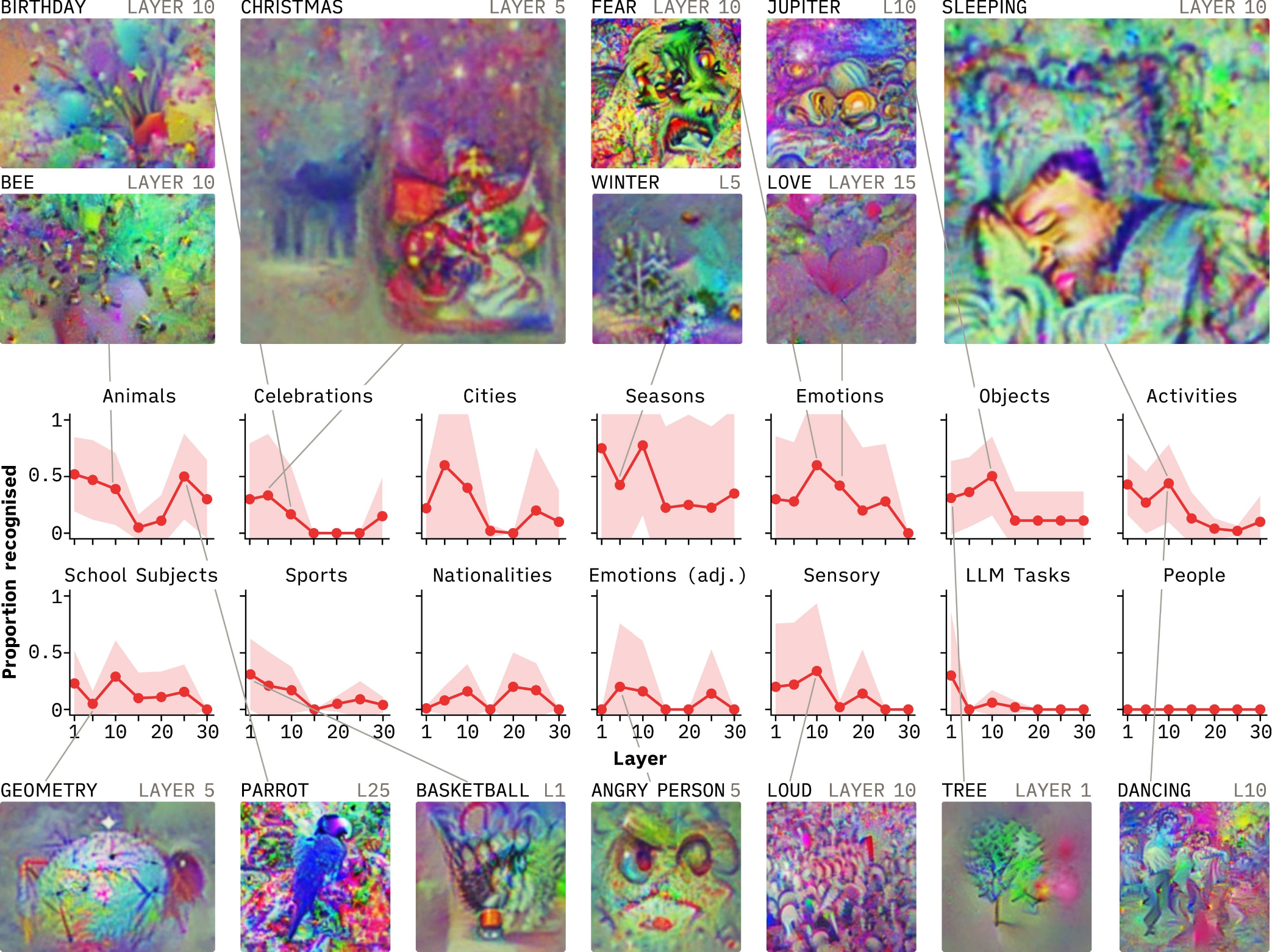}
\caption{The proportion of synthesised images that GPT-5 recognised as depicting their target concept, per category and per layer. Results shown are for evaluation with category hints; without hints, only animal concepts achieve reliable recognition. Example images show best results per category. Shaded regions: 95\% CI. Images cropped for visibility.}
\label{fig:dictionary}
\end{figure*}

To investigate when representations of concepts extracted from text and images align in VLMs, we apply our method across many concepts and layers.
Given a concept like \textit{Jupiter}, we extract its concept vector from the text ``Jupiter'' at a target layer, then synthesise an image whose representation in that layer aligns with the vector.
Then, we test whether an independent model can recognise the resulting image as depicting Jupiter.
The image will be recognisable only if the model's representations of Jupiter-as-text and Jupiter-as-image correspond at the target layer.

There is a fundamental asymmetry in our method: it constructively demonstrates alignment, but cannot prove its absence, since a failure to synthesise a recognisable image might be caused by a failure of optimisation rather by the non-existence of alignment.
This means the results below are only a lower bound on the true extent of alignment.

\subsection{Experimental Setup}\label{section:results_experimental_setup}

We tested over 100 concepts across 13 categories: animals (10), famous people (10), concrete objects (10), cities (5), nationalities (10), emotions (5), emotion adjectives (5), LLM tasks (5), seasons (4), celebrations (6), activities (10), sports (10), school subjects (10), and sensory sensations (5).
For each concept, we generated prototype images at seven layers of Gemma 3 4B~\cite{team2025gemma}: layers 1, 5, 10, 15, 20, 25, and 30 (of 34 total), optimising for 600 steps using one of two sets of hyperparameters depending on the layer. 
See \cref{appendix:results_experimental_setup} for details.
We also tested the animals category on InternVL 3 8B~\cite{zhu2025internvl3exploringadvancedtraining}.

To evaluate recognisability, we used GPT-5 with two protocols: a stringent version (``What is in the image if you had to guess? One word.'') and a more lenient version with category hints (``What [category] is in the image if you had to guess? One word.'').
For each image and each question type, we collected 10 response samples and used GPT-5-mini with a binary rubric to assess whether the image depicted the target concept. 
See \cref{appendix:results_experimental_setup} for details.
Without category hints, only animal concepts produce reliably recognisable images.
In the main text, we therefore report results that do use category hints, as this provides signal across more categories.
\Cref{appendix:results_hard_question} contains results without category hints as well as discussion of why most non-animal categories fail under the stricter protocol.

\subsection{Results: Early Alignment, Middle-Layer Collapse} \label{section:early_alignment}

\Cref{fig:dictionary} shows how often GPT-5 recognised the images synthesised on Gemma 3 4B~\cite{team2025gemma} as depicting their target concepts, by category and layer.

\paragraph{Key finding.} Multiple concept categories produce recognisable images from layer 1: more than 50~\% of animals, 70~\% of seasons, 50~\% of activities, and 30~\% of celebrations.
These results demonstrate early alignment between the representations of text and images for some concepts, challenging the prevailing view that such alignment only develops in mid-to-late layers~\cite{venhoff2025visual,neo2025interpretingvisualinformationprocessing}.
This is also surprising because Gemma is never explicitly trained to align text with images --- the model simply learns to process visual tokens alongside text for next-token prediction, with the vision encoder frozen during this training.

\paragraph{Layer-by-layer patterns.}

Early layers (1, 5, 10) show strong recognisability in animals (40--50~\%), seasons (50--75~\%), activities (40--50~\%), and emotions (30--60~\%), with most peaking at layer 1 or 5.
Cities, celebrations, objects, school subjects, sports, emotion adjectives, and sensory words show weaker but non-zero signal (around 25~\%).
Three categories fail almost entirely: famous people, nationalities, and LLM tasks produce near-zero recognisable images across all layers, suggesting these concepts either lack visual grounding or our method cannot find it.

In middle layers (15, 20), recognisability drops sharply for nearly all categories that showed early success.
This mirrors our findings on the toy experiment in \cref{appendix:motivation} and merits further investigation (see below).

Late layers (25, 30) show partial recovery in recognisability for animals, celebrations, and cities (approximately 15--25~\%), but most categories remain at zero.
No category fully recovers to its early-layer recognisability.

\paragraph{Understanding the middle-layer collapse.}

\begin{figure}
\centering
\includegraphics[width=0.8\columnwidth]{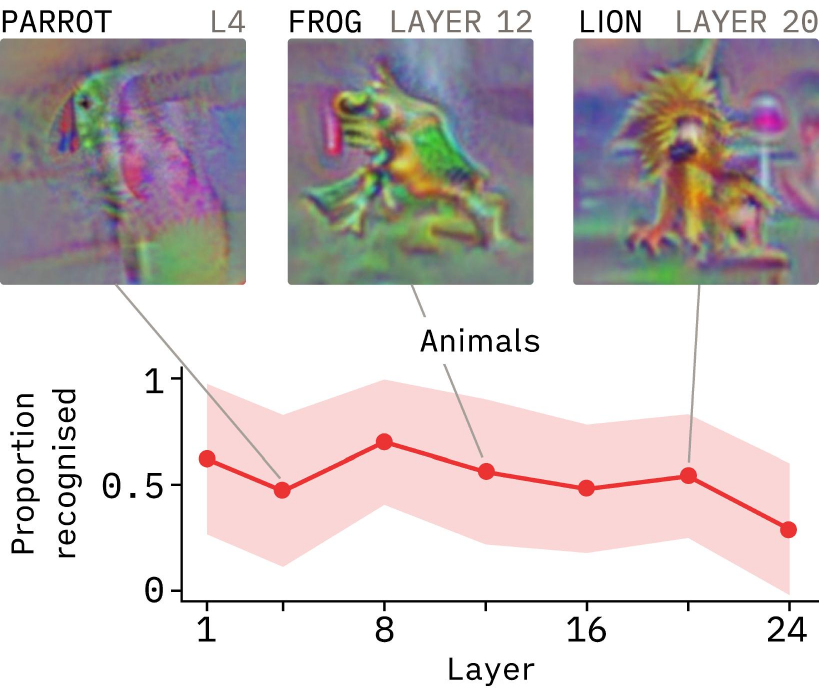}
\caption{Images generated on InternVL 3 8B maintain high recognisability even in middle layers, unlike ones generated on Gemma 3 4B where recognisability drops to near zero in layers 15--20 (\cref{fig:dictionary}).
Evaluation with category hints; example images show best results.}
\label{fig:internvl_animals}
\end{figure}

The sharp drop in layers 15--20 --- which mirrored the observations we made in \cref{appendix:motivation} on a toy experiment with two image classes --- led us to investigate possible causes through two exploratory experiments on the animals category.

First, we tested whether our semantic attention weighting contributes to the collapse.
When we aggregate patches using a simple mean instead of our attention-weighted sum, the middle-layer collapse persists, as shown in \cref{fig:gemma_mean} in \cref{appendix:results_gemma_mean}.
This suggests that the issue lies deeper in how Gemma's middle layers encode multimodal information, possibly reflecting more abstract processing, disconnected both from the early-layer ``input space'' and the late-layer ``logit space''~\cite{lad2025remarkablerobustnessllmsstages}.

Second, we tested whether this pattern is model-specific.
As shown in \cref{fig:internvl_animals}, InternVL 3 8B~\cite{zhu2025internvl3exploringadvancedtraining} exhibits no such collapse in middle layers.
This suggests the phenomenon may be specific to Gemma 3.

These two experiments suggest the middle-layer collapse may reflect how our choice of loss function interacts with middle-layer representations in Gemma.

\paragraph{Text in synthesised images.}

\begin{figure}
\centering
\includegraphics[width=0.8\columnwidth]{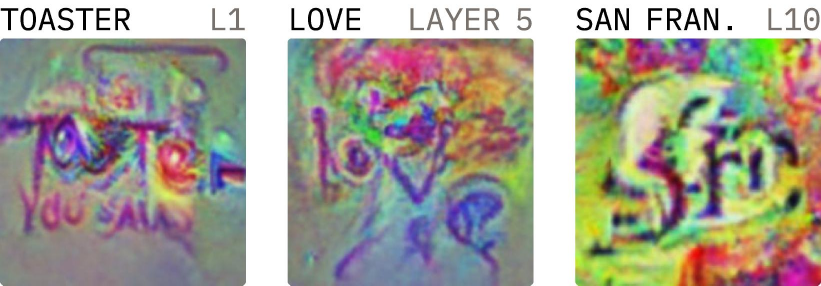}
\caption{
Some synthesised images do not visually depict the target concept but instead contain embedded text that describes the concept, suggesting the model can ``read images'' from layer 1.
}
\label{fig:text_in_image}
\end{figure}

Many synthesised images do not depict the visual representation of the target concept, but contain literal text that describes the concept, which you can see illustrated in \cref{fig:text_in_image}.

This happens from layer 1 onwards, suggesting that the model learns to associate text embedded in images with language concepts very early on, despite images and text using entirely different tokenisation schemes.

We have not quantified the split between synthesised images with and without text.
Nevertheless, our results suggest that for some concepts, the model aligns three representational forms: images depicting the concept, text describing it, and, notably, images embedding this text.
This parallels the phenomenon of multimodal neurons~\cite{goh2021multimodal}, which respond to the same concept across different input modalities.

\section{Related Work}

\paragraph{Modality gap and vision-language alignment in VLMs.}

A fundamental question in VLMs is when representations of concepts from images and text align.
Prior work identified a persistent geometric separation between image and text embeddings that they called a ``modality gap''~\cite{liang2022mindgapunderstandingmodality, jiang2024hallucinationaugmentedcontrastivelearning}, and found that alignment emerges primarily in mid-to-late layers~\cite{venhoff2025visual,neo2025interpretingvisualinformationprocessing}.
Recent studies have explored the properties of this gap~\cite{yi2025deciphermodalitygapmultimodal} and methods to reduce it~\cite{NEURIPS2024_3d007df4}.

In pure language models, sensory words have been observed to induce text representations to be more aligned with those of vision and audio encoders \cite{wang2025wordsmakelanguagemodels}. 
In VLMs,
\citet{venhoff2025visual}'s analysis provided coarse-grained evidence for mid-to-late layer alignment using proxy metrics like reconstruction error on sparse autoencoders~\cite{cunningham2023sparseautoencodershighlyinterpretable} trained on text and applied to image patch positions.
\citet{neo2025interpretingvisualinformationprocessing} used LogitLens~\cite{nostalgebraist2020logitlens} to decode image patches, demonstrating that patches containing objects can be decoded to the objects' textual descriptions. 
However, LogitLens relies on the model's unembedding matrix, making it inherently biassed toward finding patterns in later layers.
Our synthesis-based method enables direct, concept-level analysis across all layers without such architectural constraints.

\paragraph{Interpretability across modalities.}
Interpretability research has approached VLMs from both unimodal and multimodal perspectives.

Unimodal methods analyse vision and language components separately: sparse autoencoders decompose activations into interpretable features for both language models~\cite{cunningham2023sparseautoencodershighlyinterpretable, bricken2023monosemanticity} and vision transformers~\cite{abdulaal2024xrayworth15features, zaigrajew2025interpretingcliphierarchicalsparse}, while steering vectors enable behavioural control~\cite{turner2024steeringlanguagemodelsactivation, panickssery2024steeringllama2contrastive}.

Multimodal approaches include attention-based studies of information flow, showing how visual information transfers from image to text tokens in early-to-mid layers~\cite{kaduri2024whatsimagedeepdivevision, zhang2025crossmodalinformationflowmultimodal}, and multimodal sparse autoencoders that jointly decompose both modalities~\cite{lou2025saevinterpretingmultimodalmodels}.

However, none of these methods directly probes whether the representations of concepts extracted from text and images align at each layer.

\paragraph{Interpreting concepts through synthesis versus search.}
Most interpretability methods, particularly those that use sparse autoencoders, interpret discovered features by searching datasets for examples that maximally activate them~\cite{bau2017networkdissectionquantifyinginterpretability, abdulaal2024xrayworth15features, templeton2024scaling}.

Although dataset search can effectively identify patterns, it risks introducing confounders and requires careful validation to avoid spurious correlations~\cite{stevens2025sparseautoencodersscientificallyrigorous}.

The alternative tradition uses generative synthesis: DeepDream~\cite{mordvintsev2015inceptionism, olah2017feature} pioneered optimising images to maximally activate specific neurones, producing striking visualisations of learnt features and inspiring subsequent work~\cite{goh2021multimodal}.

Our method extends this tradition in two key ways: we target concept directions rather than individual neurones, optimising for representational alignment instead of activation magnitude, and we employ modern techniques from adversarial robustness, specifically Direct Ascent Synthesis~\cite{fort2025directascentsynthesisrevealing}, which uses multi-resolution optimisation and augmentations for more natural results, building on multi-scale adversarial robustness and generative attacks ~\cite{fort2024ensembleeverywheremultiscaleaggregation}.
This synthesis approach provides direct evidence of alignment without the confounders inherent in dataset search.

\section{Limitations and Future Work}\label{sec:limitations}

Several limitations restrict the broader applicability of our method.

Most critically, many synthesised images are recognised by the evaluating model (GPT-5) but would not be easily recognisable to humans.
Our method produces human-recognisable images for some concepts, but does not do so reliably across all categories.
This distinction matters: for our current application --- testing whether alignment exists --- model-based transfer is sufficient, but future applications as an interpretability tool would require human recognisability, which is a stronger requirement.

Additionally, recognition rates vary substantially across categories and evaluation protocols.
With category hints, multiple categories exceed 50\% recognition; without hints, only animals achieve reliable recognition.
We provide a detailed analysis of these differences in \cref{appendix:results_hard_question}.

We primarily studied a single model (Gemma 3 4B) with only limited validation on a different architecture (InternVL 3 8B); tests on more models are needed to establish consistent patterns.
The optimisation process is also sensitive to hyperparameters, requiring layer-specific tuning that we have not fully explored.

Despite these limitations, the method's core principle --- synthesising images whose representations align with those of textual concepts --- opens exciting future directions.
This approach could be particularly powerful for early fusion models like Chameleon~\cite{chameleonteam2025chameleonmixedmodalearlyfusionfoundation}, where visual and textual representations are inherently more integrated, potentially yielding richer cross-modal visualisations.
There are also many potential methodological improvements: examining different token positions across layers, developing better aggregation methods for image patches, and finding ways to modulate between text-in-image representations and visual depictions of concepts.
Most ambitiously, with improved reliability and additional research, this method could become a general interpretability tool: given a concept direction or a sparse autoencoder feature, it could generate a prototype image revealing the model's visual associations with that concept, providing a window into how vision-language models ground abstract concepts in the visual world.

\section{Conclusion}

We have shown that vision-language models align representations of images and text far earlier than previously thought --- as early as layer 1 for many concepts --- despite never being explicitly trained to do so.
This finding emerged from a simple insight: to investigate whether a concept like \textit{Jupiter} has aligned representations, we can directly synthesise an image optimised to match the representation of the text ``Jupiter'' and test whether this image is recognisable as depicting Jupiter.
The resulting images provide constructive evidence revealing that early cross-modal representation alignment emerges as a direct consequence of end-to-end VLM training.
The applications of our method go beyond this finding: our synthesis-based approach offers a first step toward interpretability tools that provide a direct window into models' representation spaces, free from the constraints and biases of dataset search. 
We believe that our synthesis-based approach could complement existing interpretability methods by providing visual prototypes for models' features.

\section*{Acknowledgements}

We would like to thank Adam Křivka, David Africa, Liv Gorton, Rylan Schaeffer, Seb Wilkes, Isha Gupta, Kola Ayonrinde, Jakub Kryś, and Joseph Miller, for their thoughtful comments and ideas.
Evžen would also like to thank Dominika Dobišová for moral and emotional support.

\bibliography{example_paper}
\bibliographystyle{icml2026}

\newpage
\appendix

\onecolumn

\section{Motivating Example: Comparing Apples and Oranges} \label{appendix:motivation}

Our method relies on maximising cosine similarity between image and text representations (\cref{section:concept_directions}), aggregating patch activations which discards crucial spatial information (\cref{section:patch_aggregation}), and optimising in high-dimensional spaces where most images are meaningless noise (\cref{section:das}).
Why should this produce meaningful visualisations?

These concerns deserve careful investigation.
Past work examining VLMs offers some high-level clues.
First, the linear representation hypothesis suggests that neural networks encode concepts as directions in the activation space~\cite{mikolov2013distributedrepresentationswordsphrases, cunningham2023sparseautoencodershighlyinterpretable}, making cosine similarity a natural measure of alignment.
VLMs also show surprising robustness to patch reordering~\cite{qi2025semanticsrediscoveringspatialawareness}, suggesting that aggregated patches might preserve semantic content.
Yet, this does not reveal how cosine similarity actually behaves across layers.
For that, we need empirical evidence.

We computed $\operatorname{rep}_\ell(w)$ for ``apple'' and ``orange'' at each layer using our method described in \cref{section:concept_directions}, then measured their similarity to representations of images from three sources: 100 apple images and 100 orange images~\cite{CycleGAN2017}, and 100 white noise controls.
At each layer, we computed the cosine similarity between $\operatorname{rep}_\ell(w)$ and each image's mean-aggregated patch activations, which we first centred as in \cref{section:patch_aggregation}.
We also measured the maximum patchwise similarity between the text representations and the images, and we report it in \cref{appendix:motivation_internvl}.

\Cref{fig:apple_orange} reveals that in Gemma 3 4B, cosine similarities behave in very complex ways.
Similarities fluctuate dramatically across layers, sometimes falling into the negatives; similarities for different concepts operate at different scales; and surprisingly, in middle layers (approximately 15--22), both fruit concepts are more similar to white noise than to any natural images.
Yet, a crucial pattern emerges: in early (1--15) and late (22--34) layers, text representations consistently show higher similarity to representations of their matching images. 
That is, outside the middle-layer anomaly, the representation of ``apple'' aligns more with apple image representations than orange image representations, and vice versa ($p < 0.01$, permutation test).

\begin{figure}[h]
\hspace{1em}
\centering
\includegraphics[width=0.35\columnwidth]{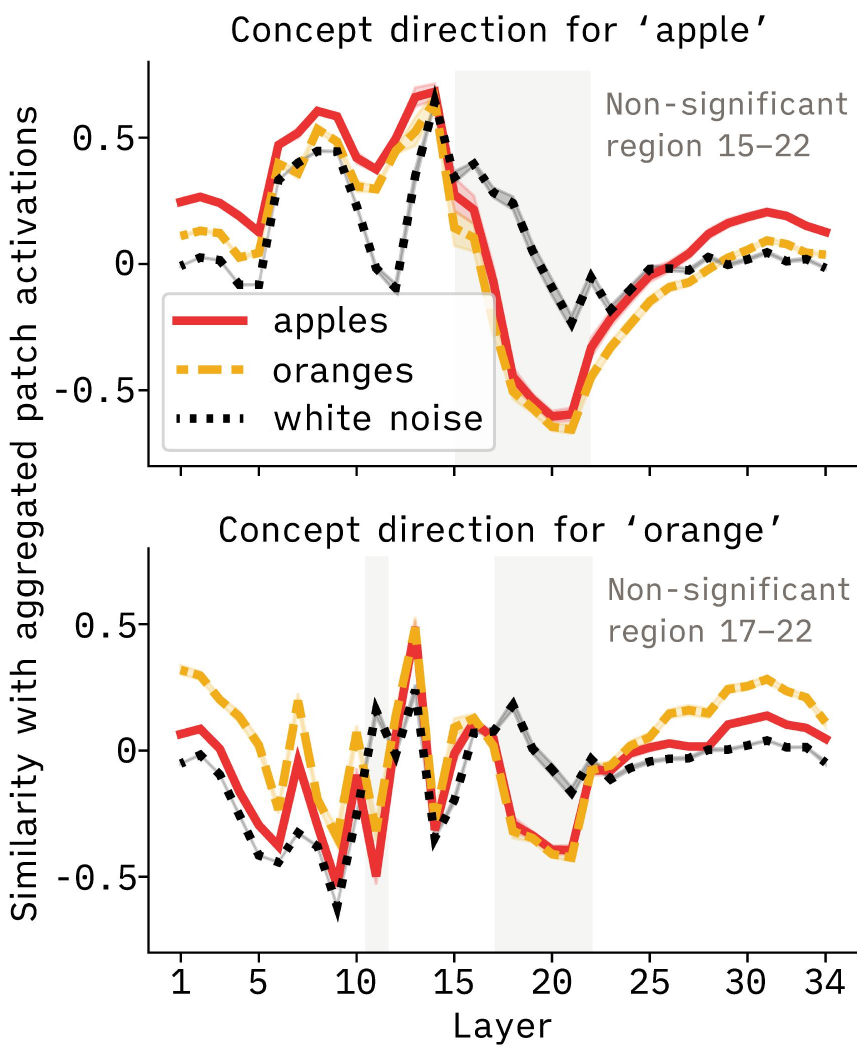}
\caption{Layer-wise cosine similarity between text representations (extracted via our method) and image representations in Gemma 3 4B. 
Despite erratic absolute values, matching concept-image pairs achieve reliably higher similarity than mismatched pairs in early and late layers ($p < 0.01$).
Each line shows mean across 100 images; shaded regions show 95\% CI (nearly invisible due to low variance).
}
\label{fig:apple_orange}
\end{figure}

The fact that similarities reliably indicate which images match which concepts in most layers suggests that choosing cosine similarity as the target for optimisation might produce meaningful results.
Specifically, by maximising similarity of an image's representation with a concept vector, we would expect to be navigated towards the class of images that genuinely represent that concept.
This also provides intuition for why dataset-based approaches that search for maximally activating examples can be meaningful, provided that the dataset contains relevant examples and is free of confounders and biases. 

To test robustness, we repeated this analysis on InternVL 3 8B~\cite{zhu2025internvl3exploringadvancedtraining}.
The positive findings replicate: one can discriminate between matching and mismatched concept-image pairs based on cosine similarity in early and late layers; see \cref{fig:apple_orange_agg_internvl} in \cref{appendix:motivation_internvl}.
Moreover, in InternVL, this is true of the middle layers, too; the middle-layer anomaly we observed in Gemma does not appear.
This suggests that our core finding (relative similarities are meaningful and provide a good target for optimisation) is robust, while specific layer-wise quirks are model-dependent.

These patterns remarkably mirror our main results in \cref{section:results}, in which we use our method to synthesise images that visually correspond to target textual concepts.
The synthesis works exactly in the layers where we saw that cosine similarity can discriminate between matching and mismatched image-concept pairs: early and late layers in Gemma, and all layers in InternVL.
Correspondingly, the synthesis fails in Gemma's middle layers (15, 20) where we saw the middle-layer anomaly occur in the results above.
This tight correspondence motivates the use of optimisation and cosine similarities to generate visual representations of concepts extracted from text.

\section{Full Formulation of the Loss Function Used for Image Synthesis}\label{appendix:full_loss_function}

The loss function we use for training is
\[
\mathcal{L}_{\operatorname{rep}_\ell(w)}(l_{448\times 448}, \ldots, l_{8 \times 8}) = -\operatorname{sim}\bigg(\operatorname{rep}_\ell(w), \operatorname{rep}_\ell\Big(\operatorname{augment}\Big(I_\text{grey} + \frac{1}{2}\operatorname{tanh}\big(\sum_r \operatorname{upscale}_{448\times 448}(l_{r \times r})\big)\Big)\Big)\bigg),
\]
where $l_{r\times r}$ are the trainable multi-resolution components we describe in \cref{section:das}, $\operatorname{upscale}_{448\times 448} : \mathbb{R}^{* \times *} \to \mathbb{R}^{448 \times 448}$ is a function that upscales any component to the target size $448 \times 448$, $I_\text{grey} \in \mathbb{R}^{448 \times 448}$ is the neutral grey image (pixels = 0.5) we use as base, $\operatorname{augment} : \mathbb{R}^{R \times R} \to \mathbb{R}^{R \times R}$ is the augmentation function that adds random noise and shifts to images we describe in \cref{section:das}, $\operatorname{rep}_\ell : \text{Image} \to \mathbb{R}^{H}$ is the function that computes the aggregate image representation as described in \cref{section:patch_aggregation}, $\operatorname{rep}_\ell(w)$ is the representation of the target textual concept at layer $\ell$ as described in \cref{section:concept_directions}, and $\operatorname{sim} : \mathbb{R}^{H} \times \mathbb{R}^{H} \to \mathbb{R}$ is the cosine similarity.

\section{Baseline Construction for Word Activations}\label{appendix:word_baseline}

In \cref{section:concept_directions}, we described that to compute the representation of a target concept $w$ in layer $\ell$, we extract its activation and subtract a baseline:
$$
\operatorname{rep}_\ell(w) = \operatorname{activation}_\ell (w) - \mathbf{b}^\text{lang}_{\ell},
$$
where $\mathbf{b}^\text{lang}_{\ell} \in \mathbb{R}^{H}$ is the baseline vector, computed as the mean activation across a set of 100 baseline words $W_\text{baseline}$:
$$
\mathbf{b}^\text{lang}_{\ell} = \frac{1}{|W_{\text{baseline}}|} \sum_{w' \in W_{\text{baseline}}} \operatorname{activation}_{\ell}(w').
$$

The following words are in $W_\text{baseline}$, taken directly from the original paper by~\citet{lindsey2025emergent}, but lowercase and put in plural form: desk, jacket, gondola, laughter, intelligence, bicycle, chair, orchestra, sand, pottery, arrowhead, jewelry, daffodil, plateau, estuary, quilt, moment, bamboo, ravine, archive, hieroglyph, star, clay, fossil, wildlife, flour, traffic, bubble, honey, geode, magnet, ribbon, zigzag, puzzle, tornado, anthill, galaxy, poverty, diamond, universe, vinegar, nebula, knowledge, marble, fog, river, scroll, silhouette, marble, cake, valley, whisper, pendulum, tower, table, glacier, whirlpool, jungle, wool, anger, rampart, flower, research, hammer, cloud, justice, dog, butterfly, needle, fortress, bonfire, skyscraper, caravan, patience, bacon, velocity, smoke, electricity, sunset, anchor, parchment, courage, statue, oxygen, time, butterfly, fabric, pasta, snowflake, mountain, echo, piano, sanctuary, abyss, air, dewdrop, garden, literature, rice, enigma.

In the equations above, $\operatorname{activation}_\ell : \text{String} \to \mathbb{R}^{H}$ computes the activation in layer $\ell$ by wrapping the input text in the model's chat template, collecting the activations on all token positions belonging directly to the input text, then averaging over them to get a single $H$-dimensional vector that represents the whole input text.

\section{Full Results on Comparing Apples and Oranges}\label{appendix:motivation_internvl}

We extended the analysis from \cref{appendix:motivation}, where we examined how concept vectors for ``apple'' and ``orange'' relate to representations of images of apples and oranges across layers. Here we present additional results using an alternative similarity metric and a second model from a different architecture family.

\paragraph{Metrics.} We computed concept-image similarities using two approaches:
\begin{itemize}
\item \textbf{Aggregate similarity} (used in main text): We average all centred patch activations of an image to obtain a single representation vector for the image in the target layer, then compute the cosine similarity with the concept vector.
\item \textbf{Maximum patchwise similarity} (new): We compute cosine similarity between the concept vector and each individual centred patch activation, then report the maximum value across all patches.
\end{itemize}

\begin{figure}[h]
\centering
\begin{subfigure}[t]{0.32\textwidth}
\centering
\includegraphics[width=\textwidth]{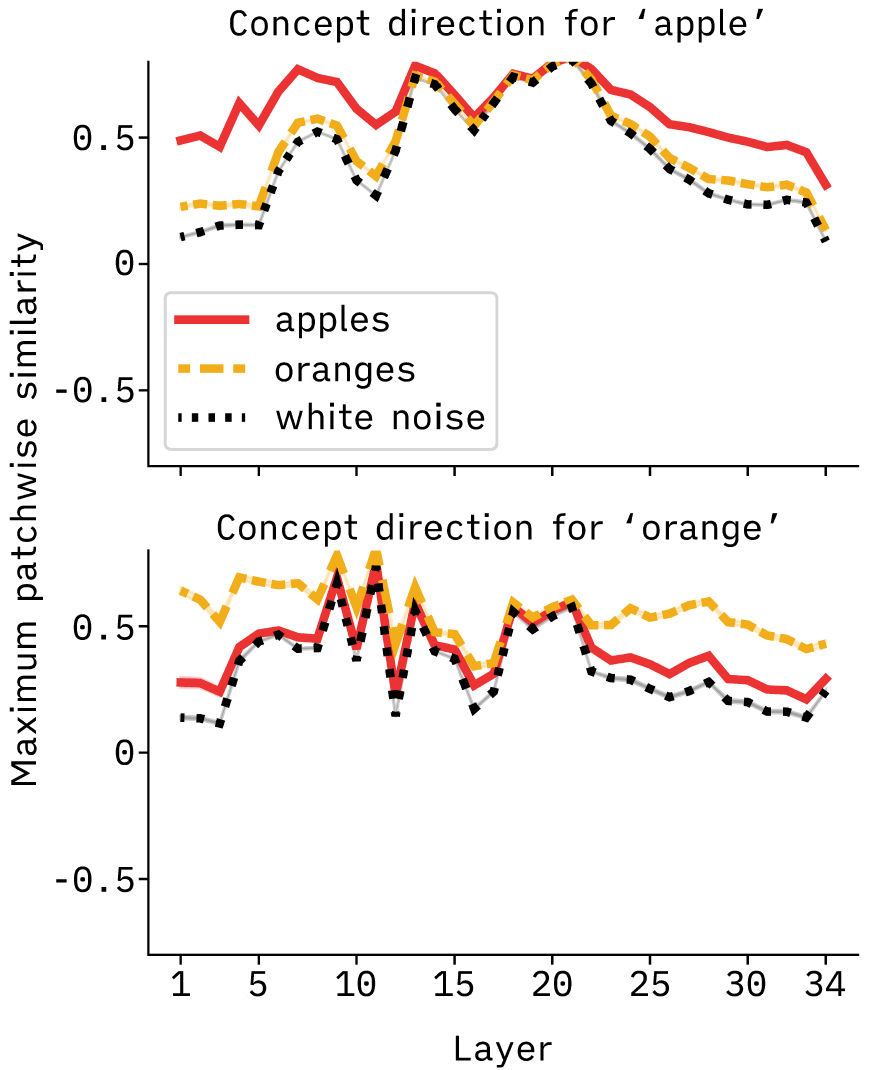}
\caption{Gemma 3 4B, maximum patchwise similarity. Matching pairs achieve higher similarity throughout ($p < 0.01$).}
\label{fig:apple_orange_max_gemma}
\end{subfigure}
\hfill
\begin{subfigure}[t]{0.32\textwidth}
\centering
\includegraphics[width=\textwidth]{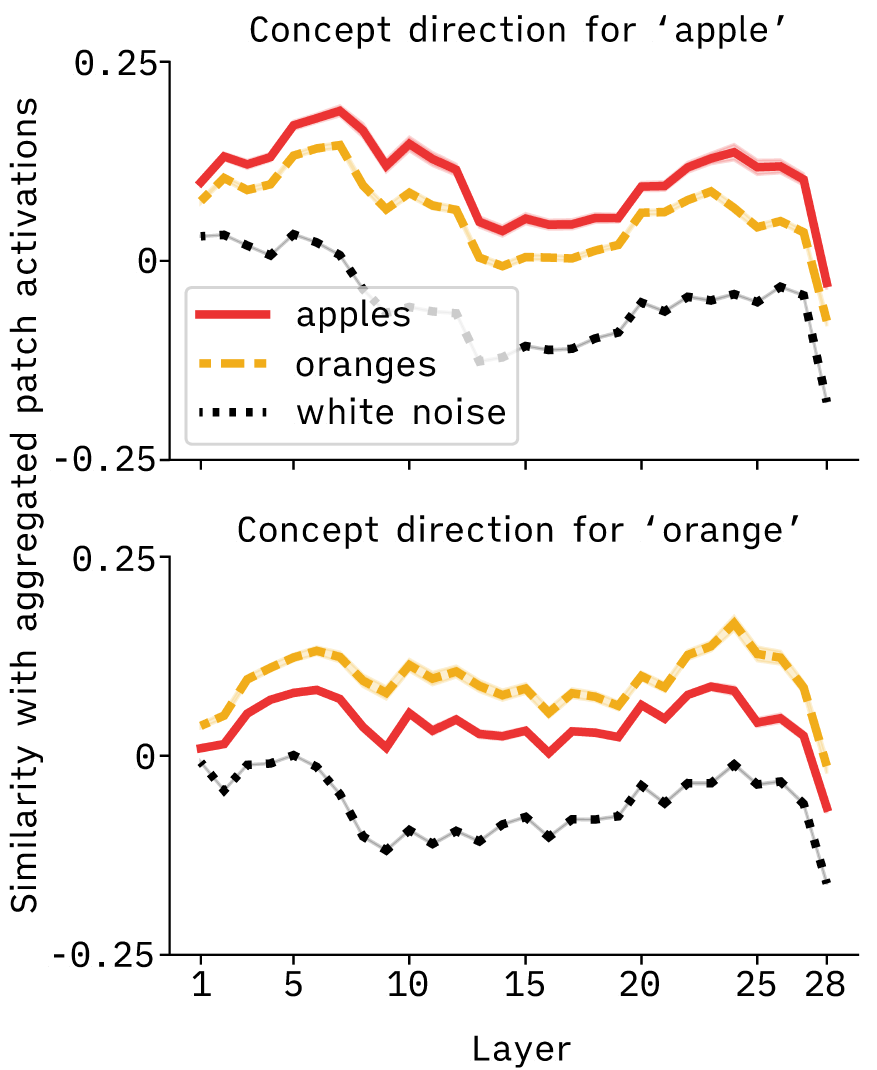}
\caption{InternVL 3 8B, aggregate similarity. No middle-layer anomaly unlike Gemma (\cref{fig:apple_orange}).}
\label{fig:apple_orange_agg_internvl}
\end{subfigure}
\hfill
\begin{subfigure}[t]{0.32\textwidth}
\centering
\includegraphics[width=\textwidth]{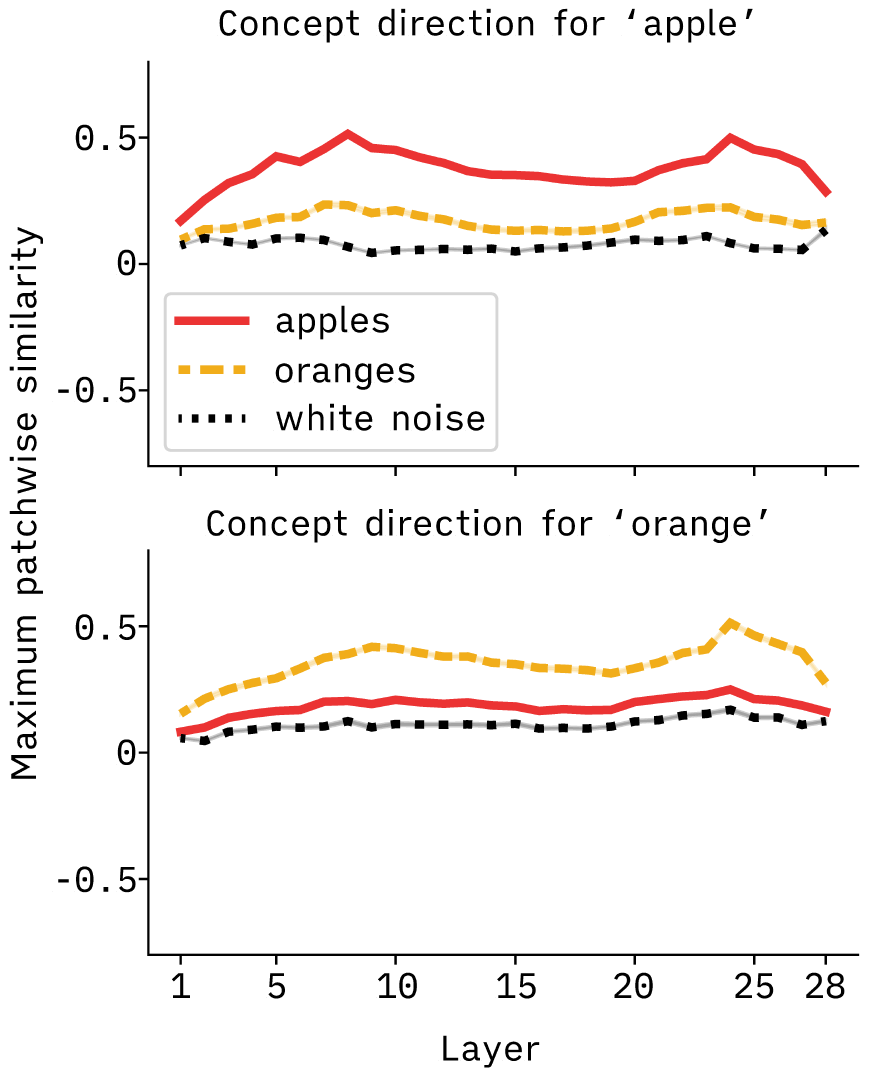}
\caption{InternVL 3 8B, maximum patchwise similarity. Consistent discrimination across all layers ($p < 0.01$).}
\label{fig:apple_orange_max_internvl}
\end{subfigure}
\caption{Cosine similarity between concept vectors (extracted from text: ``apple'', ``orange'') and image representations (100 apple images, 100 orange images, 100 white noise images) across layers in two models. Concept vectors are computed as described in \cref{section:concept_directions}. Matching pairs (apple concept with apple images, orange concept with orange images) achieve significantly higher similarity than non-matching pairs across metrics and models ($p < 0.01$, permutation test). Shaded regions show 95\% confidence intervals.}
\label{fig:apple_orange_comparison}
\end{figure}

\paragraph{Gemma results.} \Cref{fig:apple_orange_max_gemma} shows maximum patchwise similarities in Gemma 3 4B. Unlike aggregate similarity, this metric maintains discrimination between matching and non-matching image-concept pairs even in the problematic middle layers (15--22), though the gap narrows considerably there. This suggests the middle-layer anomaly specifically affects how patches combine, not individual patch representations.

\paragraph{InternVL results.} We replicated both analyses on InternVL 3 8B~\cite{zhu2025internvl3exploringadvancedtraining}. Both aggregate similarity (\cref{fig:apple_orange_agg_internvl}) and maximum patchwise similarity (\cref{fig:apple_orange_max_internvl}) show consistent discrimination between matching and non-matching pairs across all layers ($p < 0.01$, permutation test). The overall behaviour of the similarity is also notably smoother than in Gemma, lacking the erratic fluctuations and middle-layer anomaly.

\section{Full Experimental Setup for Main Results} \label{appendix:results_experimental_setup}

We tested 100+ concepts across 13 categories.

\begin{itemize}
\item \textbf{Animals:} octopus, frog, squirrel, giraffe, bee, dog, lion, elephant, parrot, T-rex
\item \textbf{Seasons:} spring (season), summer, autumn, winter
\item \textbf{Celebrations:} Christmas, Halloween, Easter, birthday, wedding, funeral
\item \textbf{Subjects:} mathematics, philosophy, geometry, history, physics, chemistry, biology, computer science, geography, music (subject)
\item \textbf{Sensory:} loud, silent, smooth, rough, sweet
\item \textbf{People:} Cleopatra, Caesar, Napoleon, Marilyn Monroe, Frida Kahlo, Elvis Presley, Einstein, William Shakespeare, Wolfgang Amadeus Mozart, Winston Churchill
\item \textbf{Cities:} new york, san francisco, paris, rome, london
\item \textbf{Sports:} soccer, poker, basketball, chess, hockey, rugby, tennis, golf, judo, boxing
\item \textbf{Nationalities:} french, italian, egyptian, czech, chinese, greek, american, indian, german, japanese
\item \textbf{Activities:} swimming, running, reading, eating, sleeping, crying, smiling, flying, screaming, dancing
\item \textbf{Emotions:} love, fear, anger, sadness, happiness
\item \textbf{Emotion adjectives:} loving person, fearful person, angry person, sad person, happy person
\item \textbf{LLM tasks:} programming, translating, refusing, summarizing, formatting
\item \textbf{Physical objects:} kettle, toaster, jupiter, armchair, bicycle, tree, camera, key, radio, phone
\end{itemize}

The optimisation uses SGD with momentum 0.9 and employs layer-dependent hyperparameters where the learning rate is 0.15 for layers 10, 15, 20, and 25, but 0.04 for layers 1, 5, 30. The training runs for 600 steps with a batch size of 8, applying gradient clipping with max norm 1.0.

The image patch aggregation uses the aggregation described in \cref{section:patch_aggregation} with a $\sigma$ that starts at $\sigma=2.0$ and increases to $\sigma=16.0$ over the training period, while $\tau$ is layer-dependent with values $\tau = 0.5$ for layers 10, 15, 20, and 25, and $\tau = 0.005$ for layers 1, 5, 30. 

To evaluate whether the image can be recognised as the concept it is meant to represent, we used GPT-5 with two protocols: a stringent open-ended question (``What is in the image? One word.'') and a more lenient version with category hints (``What [category] is in the image? One word.'').
For each image, we collected 10 responses and used GPT-5-mini with a binary rubric to assess recognition:

\begin{Verbatim}[breaklines=true]
Please carefully read the following user request (denoted as 'REQUEST') and response (denoted as 'RESPONSE').

<BEGIN REQUEST>
{request}
<END REQUEST>

<BEGIN RESPONSE>
{response}
<END RESPONSE>

You are evaluating a model that answered a question (REQUEST) about an image.

Analyze the RESPONSE, then score it according to the following rubric:

0: In the span of the answer, the model doesn't mention it sees '{concept}' in the image. E.g. it mentions other concepts, but not '{concept}'.

1: In the span of the answer, the model does mention it sees '{concept}' in the image. Either it does so explicitly, or it is clear from the context of the REQUEST and the content of the RESPONSE. It is okay if it mentions other things in addition to '{concept}'.

End your answer with "Result: 0" or "Result: 1", depending on the score.
\end{Verbatim}

\section{Full Main Results on Gemma 3 4B} \label{appendix:results_hard_question}

Here we present results using the more stringent evaluation protocol without category hints ("What is in the image? One word.") compared to the main results with hints ("What [category] is in the image? One word.").

\paragraph{Summary of main results.} With category hints (reported in main text), multiple categories produce recognisable images: animals exceed 50\% recognition in layer 1, celebrations and seasons reach 30-70\%, and activities approach 50\%. These results demonstrate that for many concepts, visual and textual representations align from the earliest layers.

\paragraph{Results without category hints.} Without hints, the evaluation becomes substantially more challenging. Animals remain the only category to consistently produce recognisable images, achieving approximately 45\% recognition in layers 1 and 5, dropping to near zero in layers 15-20, then recovering to 35-40\% in layers 25-30—mirroring the pattern observed with hints. 

A few other categories show limited recognition: objects reach 25\% in layers 1--5 and 40\% in layer 10; cities achieve 15\% in layers 5-10; activities show 15\% in layer 1 and 30\% in layer 10. Qualitative inspection reveals that successful cases in objects and cities often involve text-in-image rather than pictorial representation (such as in \cref{fig:text_in_image}). The remaining categories produce virtually no images that GPT-5 would recognise without the category hint.

\paragraph{Interpreting the gap.} The dramatic difference between evaluation protocols raises important questions about what constitutes successful synthesis and recognition. 

Regardless of the stringent no-hint results, the consistent success with animals --- and, to a lesser extent, other categories --- provides strong evidence that vision-language alignment exists for at least some concepts.
Looking at the synthesised images themselves (see \cref{fig:dictionary}), we see clear visual representations of target concepts, confirming that the alignment discovered in our main results is genuine.

Why then do most categories struggle without hints? 
We believe this reflects limitations in our synthesis method rather than absence of alignment. 
Indeed, when we provide category hints, recognition rates increase dramatically across most categories (see \cref{fig:dictionary}), suggesting the visual content is present but perhaps not crisp enough for unguided recognition.
We deliberately used consistent hyperparameters across all categories to avoid overfitting to specific concept types; targeted optimisation would likely yield clearer images and better recognition.

Moreover, even the hint-based evaluation remains highly stringent. 
For most categories, there are dozens or hundreds of possible answers (e.g., any animal, any city, any emotion), making chance-level performance negligible. The high recognition rates we observe with hints --- 50\% for animals, 50\% for activities, etc. --- cannot be explained by random guessing. They indicate genuine visual content in the synthesised images, even if that content requires context to interpret.

This remains an area for methodological improvement, as we discuss in \cref{sec:limitations}. However, the core finding stands: for multiple concept types, visual and textual representations align from the earliest layers, challenging current understanding of VLM architecture.

\begin{figure*}[h]
\includegraphics[width=0.99\textwidth]{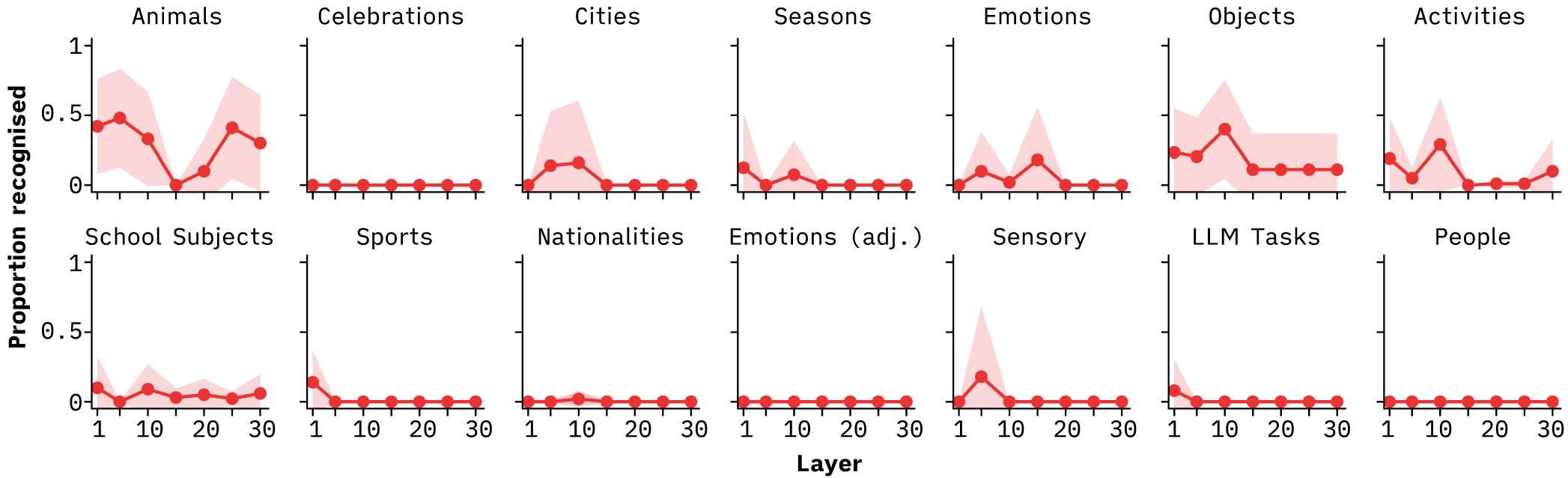}
\caption{Proportion of concepts whose prototype images GPT-5 recognised as representing the concept. Results shown are for evaluation with hard questions, without category hints. Shaded regions: 95\% CI.}
\label{fig:results_hard}
\end{figure*}

\section{Ablation: Mean Aggregation of Patches}\label{appendix:results_gemma_mean}

To test whether our attention-based patch aggregation contributes to the middle-layer collapse, we repeated the animal experiments using simple mean aggregation across all centred patch activations.

\begin{figure}[h]
\centering
\includegraphics[width=0.4\textwidth]{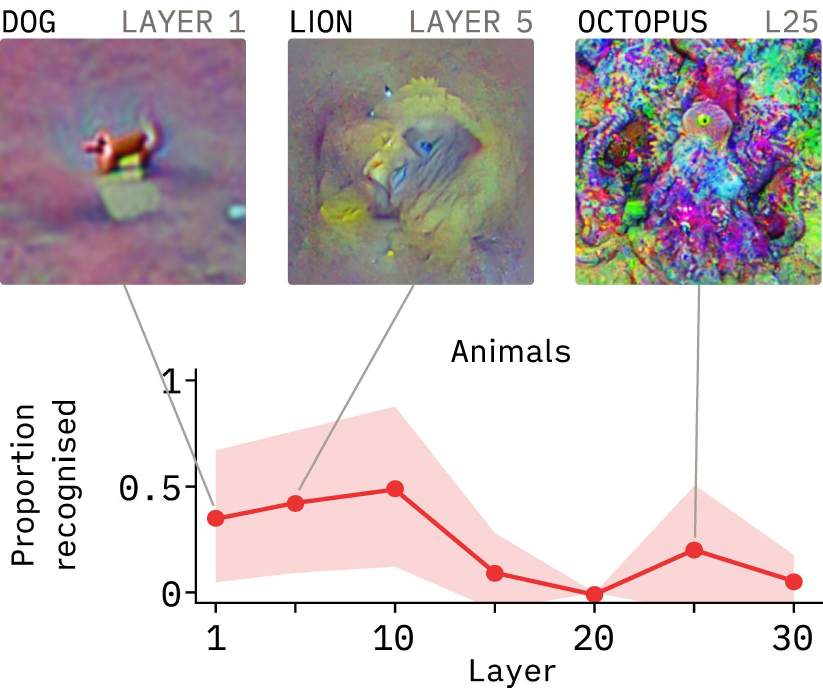}
\caption{Recognition rates for animal concepts in Gemma 3 4B using mean-aggregated patches. The middle-layer collapse persists despite the change in aggregation method. Evaluation with category hints. Shaded regions: 95\%.}
\label{fig:gemma_mean}
\end{figure}

Compared to our main results with attention-weighted aggregation (where animals achieve around 50\% in layer 1, 40\% in layer 10, near-zero in layers 15 and 20, and around 25\% in layer 25), we see that both methods suffer from the same middle-layer collapse. The key difference is that mean aggregation performs worse in layer 1 (35\% vs. over 50\%) but comparably elsewhere. Qualitatively, however, the resulting images are more diffused and unstructured.

This robustness to the aggregation method suggests that the middle-layer anomaly reflects a fundamental property of how Gemma processes multimodal information, rather than an artefact of our specific aggregation strategy.

\end{document}